\documentclass[10pt,twocolumn,letterpaper]{article}

\usepackage[pagenumbers]{cvpr} 

\usepackage{graphicx}
\usepackage{amsmath}
\usepackage{amssymb}
\usepackage{booktabs}
\usepackage[ruled,vlined,linesnumbered]{algorithm2e}
\usepackage{makecell}
\DeclareMathOperator*{\argmax}{argmax}

\usepackage[pagebackref,breaklinks,colorlinks]{hyperref}

\usepackage[capitalize]{cleveref}
\crefname{section}{Sec.}{Secs.}
\Crefname{section}{Section}{Sections}
\Crefname{table}{Table}{Tables}
\crefname{table}{Tab.}{Tabs.}

\crefname{section}{§}{§§}

\def\cvprPaperID{5418} 
\def\confName{CVPR}
\def\confYear{2022}

\begin{document}

\title{Active Learning for Point Cloud Semantic Segmentation \\ via Spatial-Structural Diversity Reasoning}

\author{Feifei Shao$^1$, Yawei Luo$^{1}$\thanks{Yawei Luo is the corresponding author} , Ping Liu$^{2}$, Jie Chen$^{1}$, Yi Yang$^{1}$, Yulei Lu$^1$, Jun Xiao$^1$ \\
$^1$ Zhejiang University, $^2$ Center for frontier AI research \\
{\tt\small sff@zju.edu.cn, yaweiluo329@gmail.com, pino.pingliu@gmail.com, chjie848@163.com} \\
{\tt\small yangyics@zju.edu.cn, luyvlei@163.com, junx@cs.zju.edu.cn}}



\maketitle

\begin{abstract}
  The expensive annotation cost is notoriously known as the main constraint for the development of the point cloud semantic segmentation technique. Active learning methods endeavor to reduce such cost by selecting and labeling only a subset of the point clouds, yet previous attempts ignore the spatial-structural diversity of the selected samples, inducing the model to select clustered candidates with similar shapes in a local area while missing other representative ones in the global environment. In this paper, we propose a new 3D region-based active learning method to tackle this problem. Dubbed SSDR-AL, our method groups the original point clouds into superpoints and incrementally selects the most informative and representative ones for label acquisition. We achieve the selection mechanism via a graph reasoning network that considers both the spatial and structural diversities of superpoints. To deploy SSDR-AL in a more practical scenario, we design a noise-aware iterative labeling strategy to confront the ``noisy annotation'' problem introduced by the previous ``dominant labeling'' strategy in superpoints. Extensive experiments on two point cloud benchmarks demonstrate the effectiveness of SSDR-AL in the semantic segmentation task. Particularly, SSDR-AL significantly outperforms the baseline method and reduces the annotation cost by up to $63.0\%$ and $24.0\%$ when achieving $90\%$ performance of fully supervised learning, respectively.

\end{abstract}
\section{Introduction}

\begin{figure}[t]
  \centering
   \includegraphics[width=0.98\linewidth]{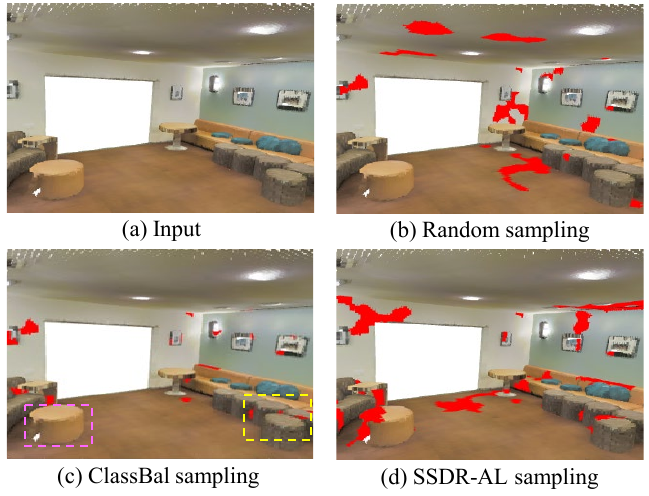}
   
  \caption{Qualitative visualization of a batch of selected samples (red regions). (a) An input point cloud. (b) Due to the random sampling strategy, the selected samples are mainly scattered in the dominating categories such as ceiling, wall, and floor. (c) ClassBal (Baseline) method balances the dominating and rare categories, but the selected regions lack spatial-structural diversity, \eg, the selected samples of chair category are all on the upper surface of the gray chair (yellow box) while ignoring the orange chair (pink box). (d) SSDR-AL (Ours) method not only considers the categorical balance but also guarantees their spatial-structural diversity.}


  \label{fig:selection}
\end{figure}

In the past few years, the field of 3D point cloud semantic segmentation has witnessed much progress with an ever-increasing number of deep learning-based methods~\cite{qi2017pointnet++,qi2017pointnet,zhao2018end,te2018rgcnn,wu2019pointconv}. The satisfying performance, however, comes with a price of expensive and laborious label annotations. Compared to its 2D counterparts~\cite{luo2019significance, luo2021category}, the annotation cost of the semantic segmentation task is more pronounced in a 3D scenario since the point clouds can be sparse, occluded, and at low resolutions. Such unbearable labeling cost makes these 3D applications infeasible at large scales~\cite{wu2021redal}.

To relieve the researchers from heavy annotation labor, some active learning-based methods~\cite{settles2008analysis,joshi2009multi,zhang2020structural,li2020hybrid,caramalau2021sequential,yuan2021multiple,cai2021revisiting} are proposed in recent years. The purpose of active learning is to incrementally select and label the most informative samples from the unlabeled data pool to reduce the overall annotation effort. In the 2D domain~\cite{luo2019taking, luo2020adversarial}, some active learning approaches~\cite{yang2015multi,yang2016active} follow an uncertainty-based sampling strategy~\cite{settles2008analysis} to select and annotate the most uncertain samples to maximize the model improvement. However, researchers~\cite{yuan2021multiple,cai2021revisiting} have gradually discovered that a simple acquisition function merely based on uncertainty is incapable of selecting the most representative samples, especially when encountering complex scenes with unbalanced categories~\cite{aggarwal2020active}. More recently, the diversity of selected samples has emerged as another important indicator in active learning sampling. For example, Cai \etal~\cite{cai2021revisiting} propose ClassBal that utilizes a class-balanced sampling strategy to increase the class-level diversity of queried samples, and Caramalau \etal~\cite{caramalau2021sequential} select the unlabeled samples that are farthest from labeled samples in latent feature space for ensuring the feature diversity of these selected samples.


Nevertheless, to adopt the active learning methods that succeed in 2D domains to the 3D point cloud data is non-trivial. The reasons are multi-fold but not limited to \textbf{1)} prior 2D region-based approaches are unable to learn the 3D spatial and structural information that reflects the representativeness of a candidate region and \textbf{2)} the internal relation between the 3D regions is rarely considered by prior works, which may cause the biased estimations on their diversity. Taking Figure~\ref{fig:selection} for example, ClassBal~\cite{cai2021revisiting} can select sample areas with more categories than the Random method since it focuses on increasing the category diversity. However, the selected samples within the same category tend to gather in a local space, which lacks spatial and structural diversities, \eg, the selected samples of the chair category are all on the upper surface of the gray chair (yellow box) while ignoring the orange chair in the pink box. In this predicament, an active learning method that enables to reason about the \emph{de facto} spatial-structural diversity and representativeness of the candidate regions are urgently needed for the point cloud semantic segmentation task. 


In this paper, we propose a new 3D region-based active learning approach, dubbed SSDR-AL, tailored for point cloud semantic segmentation via graph reasoning. SSDR-AL groups the original point clouds into superpoints~\cite{landrieu2018large,landrieu2019point} (regions) as the fundamental sample unit and projects the spatial and structural information of these superpoints into an undirected graph, as shown in Figure~\ref{fig:framework}. Specifically, we first establish an undirected graph $G=(V, E)$ in which the nodes $V=\{v_i\}$ correspond to the subset of superpoints with high uncertainty and class diversity (those low-uncertainty superpoints are dropped in this step) while the edges $E=\{e(v_i, v_j)\}$ are formed by considering both the location distance and chamfer distance between these superpoints. Then, we utilize the graph aggregation operation to merge the feature of each superpoint and its neighbors. This operation would project these superpoint features into a diversity space, in which the difference between the features in the diversity space is served as the spatial-structural distance metric. Finally, we adopt farthest point sampling (FPS) ~\cite{qi2017pointnet++,li2018pointcnn,wu2019pointconv} in such diversity space to select the most representative superpoints. In this manner, SSDR-AL enables to fully leverage both spatial-geometrical information and the internal relationship to reason about the best candidate superpoints to be labeled.

One practical problem of the superpoint-based labeling is that each superpoint may inevitably contain the points of multiple classes, which impedes the regular active learning pipeline. A fallback option to tackle this challenge is the dominant labeling method~\cite{cai2021revisiting} that treats the class of the majority points within a superpoint as the ``true label''. However, the dominant labeling strategy is prone to assign wrong labels to the minority points and result in noisy annotation. To confront this issue, we propose a simple yet effective noise-aware iterative labeling strategy. It allows the annotator to split one superpoint into a sub-region set by costing ``one click'' when the superpoint is observed to contain confounding areas. In the next, those clean sub-regions will be labeled while the confounding areas will be discarded. The proposed noise-aware iterative labeling strategy further boosts SSDR-AL in practice costing a negligible number of clicks.

The contributions of this paper can be summarised as follows:
\begin{itemize}
  \itemsep-0.1em
  \item We propose a novel superpoint-based active learning approach, dubbed SSDR-AL, for point cloud semantic segmentation. SSDR-AL can capitalize on both spatial-geometrical information and the internal relationship between superpoints to reason about the best samples to be labeled in an active learning framework.
  \item We design a noise-aware iterative labeling strategy. Being simple yet effective, it improves the previous dominant labeling-based methods which are prone to assign wrong classes to the minority points within a superpoint, which further boosts SSDR-AL in practice. 
  \item The proposed SSDR-AL achieves state-of-the-art performance on S3DIS~\cite{armeni2017joint} and Semantic3D~\cite{hackel2017semantic3d} datasets, which significantly reduces the annotation cost by up to $63.0\%$ and $24.0\%$ compared to the baseline method in achieving $90\%$ performance of fully supervised learning, respectively.
  
  \itemsep0.1em
\end{itemize}

\section{Related Work}
\subsection{Point Cloud Semantic Segmentation} 
3D point cloud semantic segmentation can be broadly categorized into two groups: graph message passing-based and neighboring feature pooling-based methods. In the first category, graph message passing-based methods consider one point cloud as a 3D graph, where the influence between points is similar to the relationship between nodes in the graph. For example, LocalSpecGCN~\cite{wang2018local} utilizes a spectral graph CNNs to establish the correlation of every local neighborhood. GACNet~\cite{wang2019graph} proposes a graph attention network to dynamically learn the structure of every object. In addition, Jiang \etal~\cite{jiang2019hierarchical} construct a hierarchical point-edge interaction network that contains a point branch and edge branch. Specifically, the point branch is responsible for predicting the class of each node, and the edge branch is designed to compute the consistency of connected nodes. In the second category, neighboring feature pooling approaches~\cite{wang2018local,wang2019graph,jiang2019hierarchical} aggregate the feature of neighboring points into a centroid point and only use these centroid points in the subsequent layers for reducing computational complexity~\cite{qi2017pointnet++,hu2020randla}. PointNet++~\cite{qi2017pointnet++} first uses the iterative farthest point sampling (FPS) to choose a set of centroid points, and then groups the neighboring points based on ball query into each centroid point. Due to the computational complexity of FPS being $O(\mathcal{N}^2)$, Randlanet~\cite{hu2020randla} replaces FPS with faster random sampling in which computational complexity is $O(1)$. To solve the segmentation performance degradation caused by the loss of key features in random sampling, Randlanet~\cite{hu2020randla} uses a local feature aggregation module to progressively increase the receptive field for each 3D point, thereby effectively preserving geometric details. Because of these merits, SSDR-AL chooses Randlanet as the segmentor of active learning.

\subsection{Active Learning}
Generally, the active learning methods can be categorized into three groups, \emph{i.e.,} uncertainty-based, diversity-based, and hybrid approaches. 

\textbf{1) Uncertainty-based Approaches.} Uncertainty indicates the difficulty of accurately predicting the unlabeled samples. These approaches~\cite{settles2008analysis,aggarwal2020active,yuan2021multiple} first leverage the model trained by labeled samples to predict the unlabeled samples, and then select those samples that are most difficult to identify. Furthermore, \cite{settles2008analysis,joshi2009multi,aggarwal2020active} groups uncertainty sampling strategies into entropy-based sampling, margin-based sampling, least confidence-based sampling and so on. Besides, MI-AOD~\cite{yuan2021multiple} computes the prediction discrepancy of two classifiers as the uncertainty of object instance.

\textbf{2) Diversity-based Approaches.} Some approaches~\cite{sener2018active,caramalau2021sequential} focus on the diversity and representativeness of the sampled data. Taking CoreGCN~\cite{caramalau2021sequential} for example, it utilizes a CoreSet technique~\cite{sener2018active} to choose the unlabeled samples that are farthest from labeled samples in latent feature space for ensuring the feature diversity of samples. 

\textbf{3) Hybrid Approaches.} The sampling function of hybrid approaches~\cite{siddiqui2020viewal,cai2021revisiting} simultaneously uses the uncertainty, diversity, and representativeness of samples. For example, ViewAL~\cite{siddiqui2020viewal} chooses the most representative superpixels that simultaneously have high entropy and low view divergence. Besides, ClassBal~\cite{cai2021revisiting} first uses the BvSB~\cite{joshi2009multi} as the uncertainty of superpixels. Then, to ensure the balanced class distribution of the selected samples, ClassBal uses a class-balance sampling strategy, which increases the weight of the tail classes while reducing it of the head classes.



Generally, SSDR-AL can be categorized as a hybrid approach since it adopts the uncertainty to initial the graph while leveraging the spatial-structural information to reason the diversity. 

\section{Methodology}

\begin{figure*}[t]
  \centering
   \includegraphics[width=1.0\linewidth]{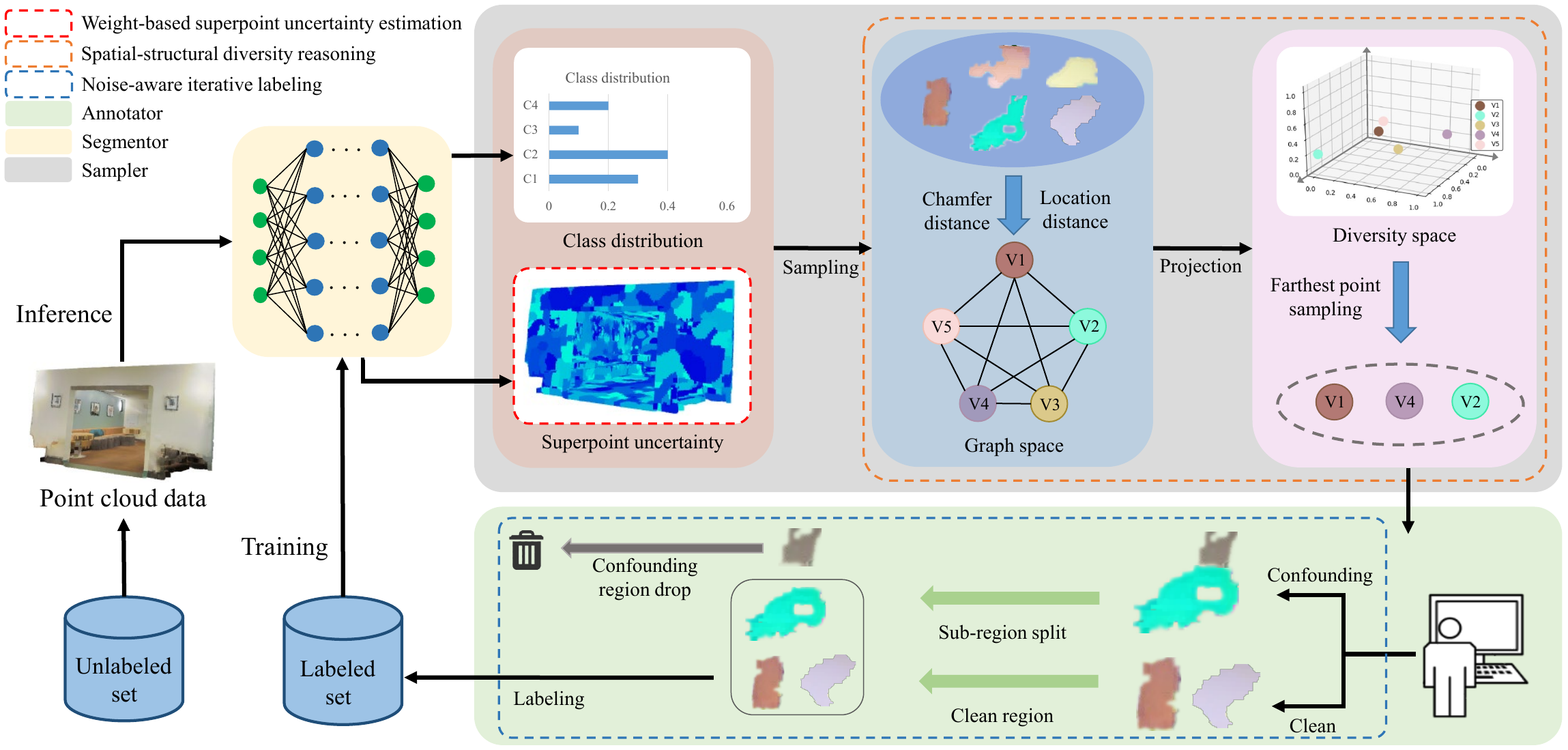}
    \caption{Overview of the proposed SSDR-AL approach. SSDR-AL loops with three steps: 1) predicting the semantic category of each point with a segmentor; 2) sampling the candidate superpoints using weight-based superpoint uncertainty estimation and spatial-structural diversity reasoning; 3) using the noise-aware iterative labeling strategy to annotate the candidate superpoints. Specifically, spatial-structural diversity reasoning consists of three parts: \romannumeral1) establishing a superpoint graph whose edges are formed by considering both the location distance and chamfer distance between superpoints; \romannumeral2) generating the features of superpoints by a graph aggregation operation to project them into a diversity space. The difference between the aggregated features in diversity space is served as the spatial-structural distance metric; \romannumeral3) selecting the most representative superpoint for annotation using farthest point sampling.}
    
   \label{fig:framework}
\end{figure*}

\subsection{Preliminaries}
\label{sec:baseline}
\textbf{Problem Settings.} Before presenting our method, we first introduce the problem settings of active learning for point cloud semantic segmentation formally. Specifically, we divide point clouds into superpoints as the fundamental sample unit of the labeled/unlabeled set. Given an annotation budget at each active learning cycle, \textbf{1)} we first select a small number of the most informative and representative superpoints from the unlabeled set and annotate them until the annotation budget is exhausted; \textbf{2)} We use the labeled superpoints to train a segmentor; \textbf{3)} Repeat 1) and 2) until the segmentor achieves the performance of $90\%$ fully supervised learning. In summary, we target costing the lowest annotation effort to achieve a certain segmentation performance by sampling and labeling the most informative and representative superpoints.

\textbf{Superpoint Generation.} Superpoint---the 3D equivalent of superpixel---has the following three properties~\cite{landrieu2019point}: \textbf{1)} each superpoint is a sub-region of one point cloud, which contains similar geometric information and must not overlap over other superpoints; \textbf{2)} the shape and contour of superpoints must be clear and coincide with the borders between objects; \textbf{3)} each point of one point cloud would only be grouped into one superpoint. In our work, we employ the global energy model~\cite{guinard2017weakly} to produce the high-quality superpoints. However, these superpoints may inevitably include noisy regions, which contain the points of multiple classes.

\textbf{Baseline Method.} Cai~\etal~\cite{cai2021revisiting} argue that a simple acquisition function based on uncertainty is incapable of querying samples from rare object categories, especially in the datasets with imbalanced class distributions. Based on such analysis, they propose the class-balanced acquisition function (ClassBal)~\cite{cai2021revisiting} that assigns different weights to the samples according to the class distribution. Specifically, they query the most informative superpixel $s^\ast$ in the unlabeled set $\mathcal{U}$ as follows:
\begin{equation}
    \label{eq:1}
  \begin{aligned}
    s^\ast = \argmax_{s \in \mathcal{U}} u(s) \ast w(d(s)),
  \end{aligned}
\end{equation}
where $u(s)$ and $d(s)$ denote the uncertainty value and dominant class (the class label of the majority points) of superpixel $s$, respectively. ClassBal regards the average uncertainty of the pixels within one superpixel as the uncertainty of this superpixel, called \textbf{Average-based Superpoint Uncertainty Estimation (AvgSU)}:
\begin{equation}
  \label{eq:avg_uncertainty}
  \begin{aligned}
    u(s) = \frac{\sum_{x \in s}u(x)}{|\{x: x \in s\}|},
  \end{aligned}
\end{equation}
where $u(x)$ denotes the uncertainty value of pixel $x$ that is a quotient of the second-best probability and the best probability~\cite{joshi2009multi}. In Equation~\ref{eq:1}, $w(\cdot)$ indicates the weight value of one class, which is given as follows:
\begin{equation}
  \begin{aligned}
  \label{eq:class_weight}
    w(c) = \exp(-\frac{|\{s: d(s)==c\}|}{|\{s: s \in (\mathcal{U} \cup \mathcal{L})\}|}),
  \end{aligned}
\end{equation}
where $\mathcal{L}$ is labeled set and $c$ denotes a certain category.

\subsection{Sampling for the Candidate Superpoints}
\label{sec:sampling}

\textbf{Weight-based Superpoint Uncertainty Estimation (WetSU).} The SOTA methods ClassBal~\cite{cai2021revisiting} and ViewAL~\cite{siddiqui2020viewal} only regard the average uncertainty of points within one superpoint as the uncertainty of this superpoint (cf. Equation~\ref{eq:avg_uncertainty}). Nevertheless, they ignore the fact that the major-class and other-class points may have variant contributions to the uncertainty of one superpoint: due to the ``one-hot'' label is usually determined by the major class, the uncertainty of other-class points actually introduces the biased estimations to the superpoint uncertainty. In other words, only the uncertainty of majority points has a positive correlation with the \emph{de facto} uncertainty of superpoint $v$. 

According to the above analysis, we first group points within one superpoint into the majority point set $\mathcal{M}$ and other point set $\mathcal{O}$ according to the model predictions. Then, we set the weight value of the majority points as $1$ and set the weight value of other points as $-1$. The uncertainty $u(v)$ of superpoint $v$ is defined as follows:
\begin{equation}
  \label{eq:weight_uncertainty}
  \begin{aligned}
    u(v) = \sum_{p_i \in \mathcal{M}}{ 1\ast u(p_i)} + \sum_{p_j \in \mathcal{O}}{(-1)\ast u(p_j)},
  \end{aligned}
\end{equation}
where $u(p_i)$ and $u(p_j)$ respectively denote the uncertainty value of point $p_i$ and point $p_j$, which is a quotient of the second-best probability and the best probability~\cite{joshi2009multi}. Finally, we use Equation~\ref{eq:1} to select the most informative superpoints but replace $u(s)$ (cf. Equation~\ref{eq:avg_uncertainty}) with $u(v)$ (cf. Equation~\ref{eq:weight_uncertainty}).



Experimentally, WetSU facilitates the purity and volume of selected superpoint $v$, both of which are preferred by active learning. Figure~\ref{fig:ablation} reports the comparison result between WetSU (cf. Equation~\ref{eq:weight_uncertainty}) and AvgSU (cf. Equation~\ref{eq:avg_uncertainty}), which verifies WetSU can select larger and more informative superpoints than AvgSU.


\textbf{Spatial-Structural Diversity Reasoning.} To select the representative superpoints with diverse geometric structures and spatial locations to enrich the sample information, we design a graph-based method to reason the spatial-structural relation between superpoints. Specifically, we establish an undirected weighted graph $G=(V, E)$ whose nodes $V=\{v_i\}$ correspond to superpoints with high uncertainty (cf. Equation~\ref{eq:weight_uncertainty}) and class diversity (cf. Equation~\ref{eq:class_weight}) from the unlabeled set, and edges $E=\{e(v_i, v_j)\}$ are built by considering both location distance and chamfer distance between these superpoints. 
On one hand, the feature $f_{v_i}$ of superpoint $v_i$ is calculated as follows:
\begin{equation}
  \begin{aligned}
    f_{v_i} = \frac{\sum_{p \in \mathcal{M}_{v_i}}{f_p}}{|\mathcal{M}_{v_i}|},
  \end{aligned}
\end{equation}
where $f_p$ is the feature of point $p$ provided by the segmentor (\eg, Randlanet in this paper) and $\mathcal{M}_{v_i}$ indicates the majority point set within the superpoint $v_i$. On the other hand, the weight value $\delta(v_i, v_j)$ of the edge $e(v_i, v_j)$ is calculated as follows:
\begin{equation}
  \begin{aligned}
    \delta(v_i, v_j) = \exp(-(D_{l}(v_i, v_j) + D_{c}(v_i, v_j))),
  \end{aligned}
\end{equation}
where location distance $D_{l}(v_i, v_j)$ is realized as Euclidean distance between the centroids of superpoints $v_i$ and $v_j$. And Chamfer distance $D_{c}$ is usually employed to evaluate the discrepancies of structure between two superpoints~\cite{mandikal2019dense,li2020through}, which is defined as follows:
\begin{equation}
  \label{eq:dis_end}
  \begin{aligned}
    D_{c}(v_i, v_j) &= \frac{1}{|v_i|}\sum_{p_a\in v_i}\min_{p_b\in v_j}||p_a-p_b||^2_2 \\ 
    &+ \frac{1}{|v_j|}\sum_{p_b\in v_j}\min_{p_a\in v_i}||p_b-p_a||^2_2,
  \end{aligned}
\end{equation}
where $p_a$ and $p_b$ are the points within superpoint $v_i$ and superpoint $v_j$, respectively. $||p_a-p_b||_2$ denotes Euclidean distance between point $p_a$ and point $p_b$. To reason the superpoint diversity in the context of graph $G$, we adopt a weighted sum aggregator to aggregate the feature of each superpoint and its one-hop neighbors into the target feature. Recall that the edges of graph $G$ are built from both the location distance and chamfer distance between superpoints, this aggregation operation would project these superpoint features into a diversity space in which the difference between the features in the diversity space is served as the spatial-structural distance metric. The graph aggregation operation shown in Figure~\ref{fig:framework} is calculated as follows:
\begin{equation}
  \label{eq:dis_begin}
  \begin{aligned}
    \hat{f}_{v_i} = \sum_{v_j \in \mathcal{N}_{v_i}}\delta(v_i, v_j)\ast f_{v_j},
  \end{aligned}
\end{equation}
where $\mathcal{N}_{v_i}$ and $\hat{f}_{v_i}$ denote the neighboring superpoint set of superpoint $v_i$ (including itself) and the aggregated feature of superpoint $v_i$ in the diversity space, respectively. 



Through Equation~\ref{eq:dis_begin}, we can project superpoints from a graph space into a diversity space to reason their spatial-structural diversity: if two superpoints (\eg, $v_1$, $v_5$) are close under location and chamfer metric as well as they have the overlapping neighboring superpoints in the graph space, their features in the diversity space would be similar (cf. Figure~\ref{fig:framework}). Thus, to prevent sampling multiple superpoints from a local cluster and ensure the spatial and structural diversities of selected superpoints, we utilize farthest point sampling (FPS)~\cite{qi2017pointnet++,li2018pointcnn,wu2019pointconv} in diversity space to select the most representative candidate superpoints for label acquisition. Finally, Figure~\ref{fig:selection} vividly shows the advantages of the spatial-structural diversity reasoning in sampling superpoints compared to the ClassBal~\cite{cai2021revisiting} (Baseline) method.



\subsection{Labeling for the Candidate Superpoints}
\label{sec:labeling}
In this section, we first introduce a click-based annotation cost measurement. Then, we propose a novel labeling strategy based on click cost that effectively mitigates the noise problem introduced by traditional dominant labeling~\cite{cai2021revisiting}. 

\textbf{Annotation Cost Measurement.} In the experimental environment, researchers adopt the ground truth annotation of the samples to simulate the annotation from the oracle~\cite{siddiqui2020viewal}. Due to the difficulty in comparing time and money costs, previous works propose to use the number of labeled points as a substitute for the real annotation cost~\cite{cai2021revisiting}. More recently, with the development of region-based approaches~\cite{cortes2019region}, some approaches~\cite{mackowiak2018cereals,CollingRGR21,cai2021revisiting} advocate adopting the click number instead of the number of labeled points as a more realistic measurement of annotation cost.

In this work, we consider that a one-click operation can assign a semantic category to one superpoint based on the dominant labeling strategy~\cite{cai2021revisiting}. Thus, we use the number of clicks as the annotation cost in the labeling phase of active learning and the methods are fairly compared using the same annotation budget.

\textbf{Noise-aware Iterative Labeling.} In practice, one superpoint may inevitably contain the points of multiple classes. The existing dominant labeling strategy~\cite{cai2021revisiting} regards the class of majority points within one superpoint as the class of this superpoint, which may result in assigning wrong categories to the minority points. To confront the issue, we propose a simple yet effective noise-aware iterative labeling strategy. Specifically, we first define the purity $\varphi(v)$ of a superpoint $v$ as the percentage of majority points within the superpoint $v$. If the purity $\varphi(v)$ of superpoint $v$ is lower than threshold $\theta$, we empower the oracle to consume ``one click'' to split $v$ into a sub-region set $\mathcal{R}$ according to the pseudo prediction class of points. Then, we compute the purity $\varphi(r)$ of sub-region $r \in \mathcal{R}$ and annotate it when its purity is higher than threshold $\theta$. Otherwise, sub-region $r$ will be discarded and not returned to the unlabeled set or added to the labeled set. An overview of the noise-aware iterative labeling strategy is provided in Algorithm~\ref{Alg:labeling}. Note that sometimes the noise-aware iterative labeling strategy may cause one noisy superpoint to consume multiple clicks, which causes some candidate superpoints to be returned to the unlabeled set because the click budget is exhausted.

\begin{algorithm}[t]
  \KwIn{Unlabeled superpoint set $\mathcal{U}_{t}$, labeled superpoint set $\mathcal{L}_{t-1}$, candidate superpoint set $\mathcal{C}_{t}$ selected by sampler (cf. Sec~\ref{sec:sampling}), annotation click budget $K_t$ for batch $t$, purity threshold $\theta$}
  \KwOut{Unlabeled set $\mathcal{U}_{t+1}$, labeled set $\mathcal{L}_{t}$}
  \textbf{initialize:} $click=0$ \\
  \For{$v \in \mathcal{C}_{t}$}{
    \If{$click < K_t$}{
      Compute the purity $\varphi(v)$ of superpoint $v$\\
      \eIf{$\varphi(v) \geq \theta$}{
        Annotate $v$ and $click = click + 1$ \\
        $\mathcal{L}_{t-1} = \mathcal{L}_{t-1} \cup v$ \\
      }{Split $v$ into sub-region set $\mathcal{R}$ and $click = click + 1$\\
        \For{$r \in \mathcal{R}$}{
          \If{$\varphi(r) \geq \theta$}
          {
            Annotate $r$ and $click = click + 1$ \\
            $\mathcal{L}_{t-1} = \mathcal{L}_{t-1} \cup r$ \\
          }
        }
      }
      $\mathcal{U}_{t} = \mathcal{U}_{t} \backslash v$
    }
  }
  $\mathcal{U}_{t+1} = \mathcal{U}_{t}$ and 
  $\mathcal{L}_{t} = \mathcal{L}_{t-1}$
  \caption{Batch-Mode Noise-aware Iterative Labeling}
  \label{Alg:labeling}
\end{algorithm}

\section{Experiments}

\subsection{Datasets and Implementation Details}
\label{sec:dataset_and_details}

\textbf{Datasets.} The proposed SSDR-AL is evaluated on two public datasets, \ie, \textbf{S3DIS}~\cite{armeni2017joint} and \textbf{Semantic3D}~\cite{hackel2017semantic3d}. 1) \textbf{S3DIS} contains $271$ rooms divided into $6$ large areas~\cite{hu2020randla} and each room corresponds to a point cloud containing medium-sized 3D points. Besides, S3DIS has $10$ room types and $13$ point categories that can be adapted to most point cloud semantic segmentation tasks. All of our tests are conducted on Area $5$ validation set. 2). To verify the strong generalization of SSDR-AL, we also conduct experiments on the \textbf{Semantic3D} dataset, which contains $15$ point clouds and $8$ classes for training. Compared with S3DIS, the point number of each point cloud of Semantic3D has up to $10^8$ points, which is much bigger than the point clouds of S3DIS. All the experiments are conducted on the validation set, which is separated from the training set. 

\addtolength{\tabcolsep}{3pt}
\begin{table}[t]
  \centering
  \caption{Comparing annotation cost (click number) required to achieve $90\%$ accuracy on S3DIS and Semantic3D datasets for different active learning methods.}
  
  \label{tab:click}
  \begin{tabular}{lcc}
  \toprule
  Methods    & S3DIS & Semantic3D \\
  
  \midrule
  Random        &111.0k     & 15.5k  \\
  Entropy~\cite{joshi2009multi}   &162.2k   & 17.8k       \\
  BvSB~\cite{joshi2009multi}       &142.5k   & 14.2k    \\
  ClassBal~\cite{cai2021revisiting} (Baseline)  &70.3k  &    11.4k       \\ 
  SSDR-AL (Ours)      &\textbf{26.0k} & \textbf{8.7k}               \\ 
  \bottomrule
  \end{tabular}
\end{table}
\addtolength{\tabcolsep}{-3pt}

\textbf{Implementation Details.} We adopt Randlanet~\cite{hu2020randla} as the segmentor of active learning and use the global energy model~\cite{guinard2017weakly} to split the original point clouds into superpoints. We produce $456,764$ superpoints and randomly select $0.5\%$ superpoints with labels as seed samples to initialize the labeled set on S3DIS~\cite{armeni2017joint} dataset (produce $456,764$ superpoints and randomly select $0.8\%$ superpoints with labels as seed samples on Semantic3D~\cite{hackel2017semantic3d} dataset). In each active learning cycle, we first use Adam~\cite{2014Adam} to optimize Randlanet with $\beta_1=0.9$ and $\beta_2=0.999$ on labeled set. Randlanet has trained $30$ epochs with the initial learning rate $0.01$ that decreases by $16\%$ after each epoch on S3DIS dataset (has trained $50$ epochs with the initial learning rate $0.01$ that decreases by $8\%$ after each epoch on Semantic3D dataset). The other hyper-parameters are consistent with default values~\cite{hu2020randla}. On S3DIS dataset, we select $10$k ($3$k on Semantic3D dataset) the most informative and representative candidate superpoints from the unlabeled set and then utilize a noise-aware iterative labeling strategy with the threshold $\theta=0.9$ to annotate them until the $10$k ($3$k on Semantic3D dataset) click budget is exhausted. These labeled superpoints and sub-regions will be added to the labeled set. Totally or partially labeled superpoints will be deleted from the unlabeled set. Meanwhile, those unlabeled candidate superpoints will be returned to the unlabeled set. In the spatial-structural diversity reasoning module, we regard all nodes as aggregation nodes and project superpoints from a graph space into a diversity space by using once aggregation.

\subsection{Comparison with State-of-the-Art Methods }
\label{sec:comparison}

\addtolength{\tabcolsep}{-5pt}  
\begin{table}[t]
  \centering
  \caption{Comparing the percentage of labeled points required to achieve $90\%$ accuracy on S3DIS dataset for different active learning methods. $\diamond$ is the results in the original paper~\cite{wu2021redal}.}
  

  \label{tab:points}
  \begin{tabular}{llc}
  \toprule
  Methods    & Segmentors & Labeled points   \\
  \midrule
  ReDAL$^\diamond$~\cite{wu2021redal}     & SPVCNN~\cite{tang2020searching} & 13\%  \\
  ReDAL$^\diamond$~\cite{wu2021redal}     & MinkowskiNet~\cite{choy20194d} & 15\%   \\
  \hline
  Random      & Randlanet~\cite{hu2020randla}      &  40.9\%        \\
  Entropy~\cite{joshi2009multi}           & Randlanet~\cite{hu2020randla}   &   46.7\%    \\
  BvSB~\cite{joshi2009multi}            & Randlanet~\cite{hu2020randla}    &     43.0\%      \\
  ClassBal~\cite{cai2021revisiting} (Baseline)    &  Randlanet~\cite{hu2020randla}   &      13.3\%   \\ 
  SSDR-AL (Ours)    & Randlanet~\cite{hu2020randla}     &  \textbf{11.7\%}              \\ 

  \bottomrule
  \end{tabular}
\end{table}
\addtolength{\tabcolsep}{5pt}

\begin{figure}[t]
  \centering
  \includegraphics[width=1.0\linewidth]{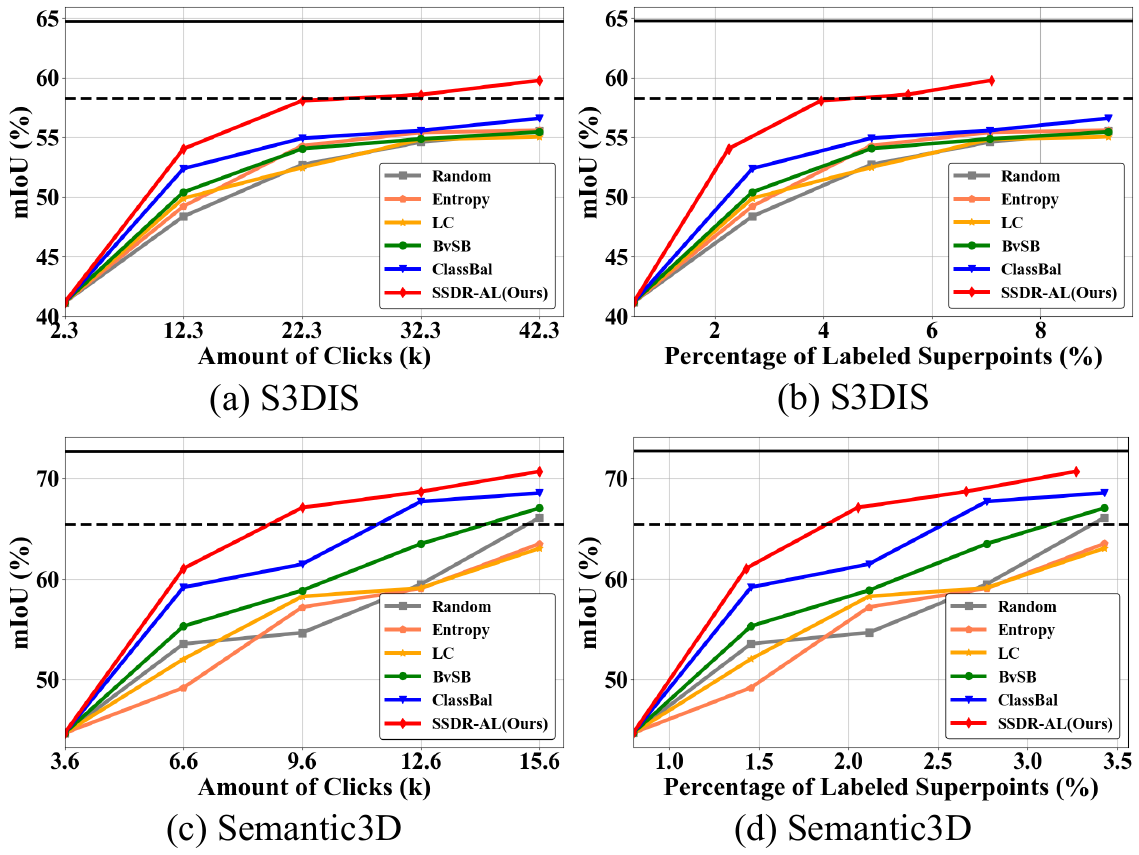}
   \caption{Comparing active learning performances of different methods on both S3DIS (a)(b) and Semantic3D (c)(d). (a)(c) Benchmarking at a fixed amount of annotation budget measured by clicks. (b)(d) Plot the same IoU results as (a)(c) but measured by the percentage of labeled superpoints. LC and BvSB denote least confidence~\cite{settles2008analysis} and  Best-versus-Second Best margin~\cite{joshi2009multi}, respectively. The uppermost solid black line and dotted line denote the maximum and 90\% performance of fully supervised learning, respectively.}
   \label{fig:sota}
\end{figure}

\begin{figure*}[t]
  \centering
  \includegraphics[width=0.98\linewidth]{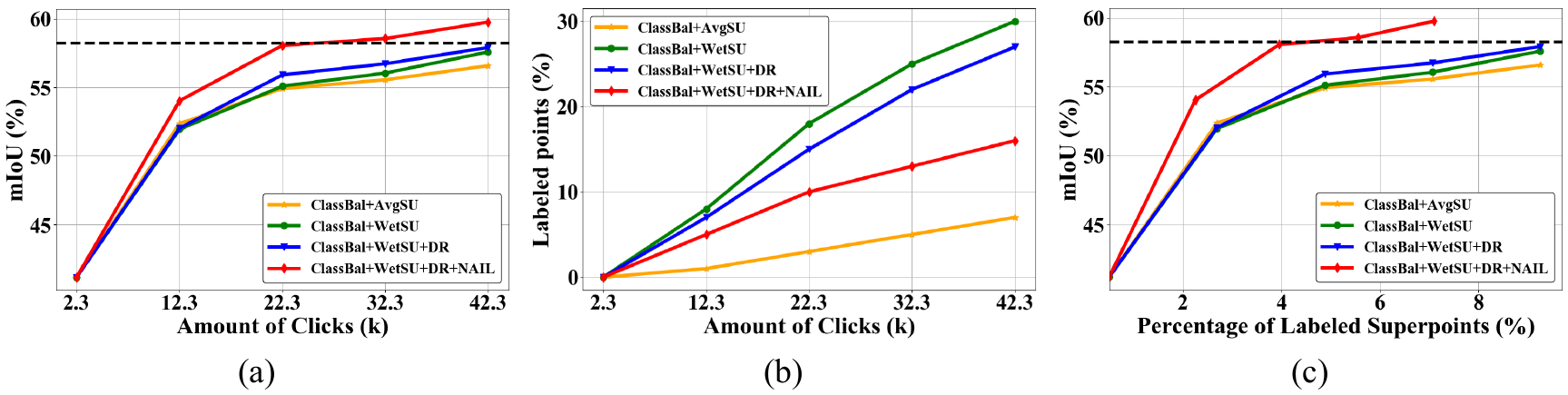}

  \caption{Ablation studies on S3DIS Area 5 validation set. (a) Benchmarking at a fixed amount of annotation budget measured in clicks. (b) Plot the percentage of labeled points at a fixed amount of annotation budget measured in clicks. (c) Plot the same results as (a) but measured in the percentage of labeled superpoints. AvgSU, WetSU, DR, and NAIL denote average-based superpoint uncertainty estimation, weight-based superpoint uncertainty estimation, spatial-structural diversity reasoning, and noise-aware iterative labeling, respectively. The uppermost dotted line denotes the $90\%$ performance of fully supervised learning.}


   \label{fig:ablation}
\end{figure*}

\begin{figure*}[t]
  \centering
  \includegraphics[width=0.98\linewidth]{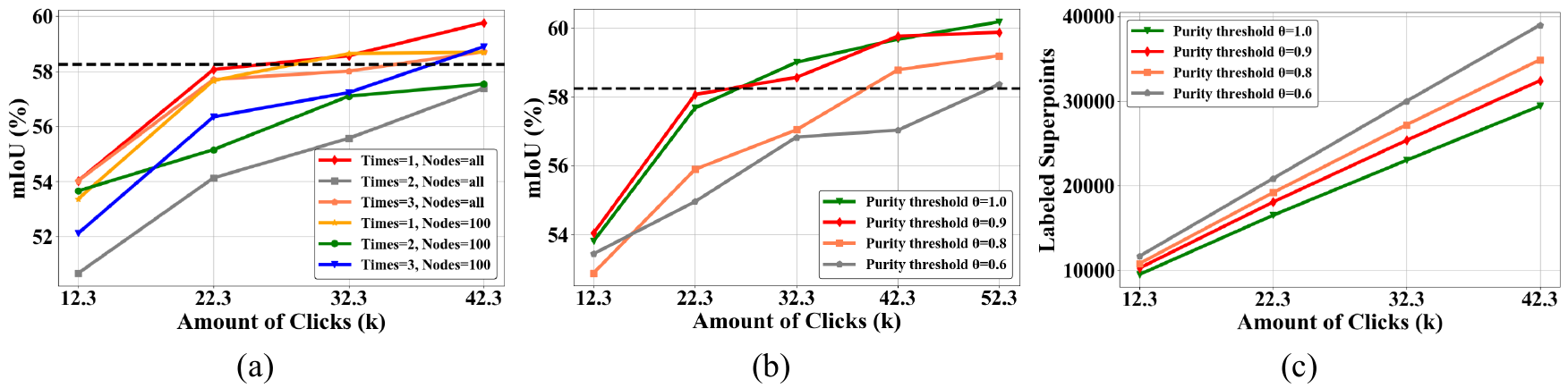}

   \caption{Analysis experiments on S3DIS Area 5 validation set. (a) Comparing the effects of the aggregation times and the number of aggregation nodes on segmentation performance. (b) Comparing the effects of different purity threshold $\theta$ of noise-aware iterative labeling strategy on segmentation performance. (c) Comparing the effects of different purity threshold $\theta$ of noise-aware iterative labeling strategy on labeled superpoints. The uppermost dotted line denotes the $90\%$ performance of fully supervised learning.}

   \label{fig:analysis}
\end{figure*}



\begin{figure*}[t]
  \centering
   \includegraphics[width=1.0\linewidth]{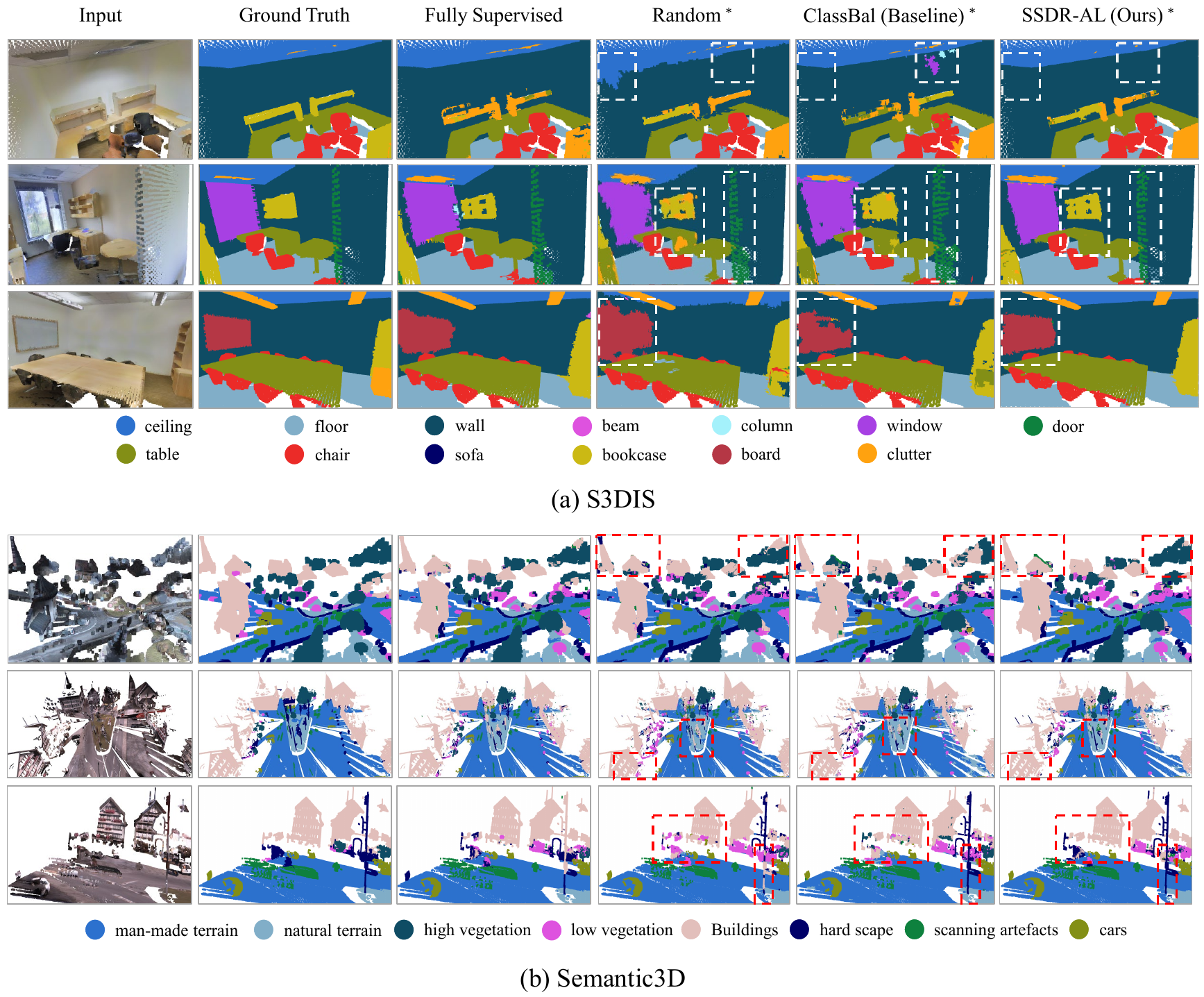}

   \caption{Qualitative segmentation results of SSDR-AL (Ours) compared with the Random, ClassBal (Baseline) on both S3DIS and Semantic3D validation sets. * indicates the segmentor trained on the final labeled set, which costs $42.3$k and $15.6$k clicks for labeling on S3DIS and Semantic3D datasets, respectively. The dotted white and red boxes indicate that our method can accurately recognize areas that other methods fail to.}
   \label{fig:representation}
\end{figure*}

We compare SSDR-AL with other state-of-the-art methods on S3DIS~\cite{armeni2017joint} and Semantic3D~\cite{hackel2017semantic3d} datasets shown in Figure~\ref{fig:sota}. Meanwhile, to gain more insights into the advantage of SSDR-AL, we plot the sampling and segmentation results from different active learning methods, which are shown in Figure~\ref{fig:selection} and Figure~\ref{fig:representation}, respectively. 

\textbf{S3DIS.} From these observations in Figure~\ref{fig:sota} (a), we conclude the findings that our SSDR-AL significantly outperforms other methods at a fixed amount of annotation budget measured in clicks, especially it has been obviously in the leading position in every active learning cycle and first reaches the performance of $90\%$ fully supervised learning \footnote{The mIoU performance of Randlanet~\cite{hu2020randla} based on fully supervised learning is $64.72\%$ on S3DIS dataset.} costing only $26.0$k clicks. Table~\ref{tab:click} reports that SSDR-AL reduces the annotation cost by up to $63.0\%$ lower compared to ClassBal~\cite{cai2021revisiting} (Baseline) method in achieving the $90\%$ performance of fully supervised learning. To validate the informative and representative quality of our selected superpoints, we provide the segmentation performance measured in the percentage of labeled superpoints in Figure~\ref{fig:sota} (b). From Figure~\ref{fig:sota} (b), we find that SSDR-AL can achieve better segmentation performance while using fewer labeled superpoints. It is worth noting that the red line stops early because SSDR-AL discards many noisy superpoints and confounding areas by using the noise-aware iterative labeling strategy. Besides, Table~\ref{tab:points} reports that SSDR-AL annotates $1.3\%$ fewer points than ReDAL~\cite{wu2021redal} in achieving the $90\%$ performance of fully supervised learning. This directly illustrates that these superpoints selected by SSDR-AL are more informative and representative than ReDAL~\cite{wu2021redal}. 





\textbf{Semantic3D.} In Figure~\ref{fig:sota} (c), we observe that our SSDR-AL significantly outperforms other methods. SSDR-AL first reaches the performance of $90\%$ fully supervised learning \footnote{The mIoU performance of Randlanet~\cite{hu2020randla} based on fully supervised learning is $72.70\%$ on Semantic3D dataset.} costing only $8.7$k annotation click. A report provided in Table~\ref{tab:click} shows that SSDR-AL reduces the annotation cost by up to $24.0\%$ lower compared to ClassBal~\cite{cai2021revisiting} (Baseline) method. Besides, Figure~\ref{fig:sota} (d) shows that at the same percentage of labeled superpoints, the segmentor (Randlanet~\cite{hu2020randla}) can achieve better segmentation performance on the training data selected by our method, which verifies the generalization and effectiveness of SSDR-AL in multiple point cloud scenarios.


\subsection{Ablation Study}
\label{sec:ablation}
To better understand the effectiveness of the weight-based superpoint uncertainty estimation, spatial-structural diversity reasoning, and noise-aware iterative labeling, we conduct several ablation studies on S3DIS~\cite{armeni2017joint} dataset. The results of our ablation studies are illustrated in Figure~\ref{fig:ablation}. We observe that the weight-based superpoint uncertainty estimation brings an extra $1.0\%$ mIoU improvement at the $4^{th}$ active learning cycle in Figure~\ref{fig:ablation} (a). Besides, weight-based superpoint uncertainty estimation (cf. Equation~\ref{eq:weight_uncertainty}) can label more points than the average-based superpoint uncertainty estimation (cf. Equation~\ref{eq:avg_uncertainty}) when costing the same annotation budget shown in Figure~\ref{fig:ablation} (b). This verifies that weight-based superpoint uncertainty estimation is prone to select large and informative superpoints. Employing weight-based superpoint uncertainty estimation and spatial-structural diversity reasoning together improves $1.2\%$ mIoU over the baseline and reaches $57.9\%$ mIoU at the $4^{th}$ active learning cycle. After further employing the noise-aware iterative labeling strategy to mitigate the noise problem introduced by dominant labeling~\cite{cai2021revisiting}, our final model reaches $59.8\%$ mIoU, outperforming ClassBal~\cite{cai2021revisiting} (Baseline) method by $3.2\%$. Besides, comparing the red and blue lines in Figure~\ref{fig:ablation} (c) shows that the noise-aware iterative labeling strategy annotates fewer superpoints than the traditional dominant labeling~\cite{cai2021revisiting} in every active learning cycle. The reason for this phenomenon is the former discards some noisy superpoints and confounding areas to alleviate their negative effects for achieving better segmentation performance. This further underscores the importance of using the noise-aware iterative labeling strategy.

\subsection{Analysis}
\label{sec:analysis}
As shown in Figure~\ref{fig:analysis} (a), we conduct several experiments to illustrate the different effects of the aggregation times and the number of aggregation nodes on segmentation performance. Besides, Figure~\ref{fig:analysis} (b) reports the effects of different purity threshold $\theta$ of noise-aware iterative labeling strategy on segmentation performance.

\textbf{Spatial-structural Diversity Reasoning.} We set up two groups of comparison experiments in Figure~\ref{fig:analysis} (a): the same number of nodes and the same number of aggregations. In the case of the same number of aggregation nodes, we observe that once aggregation is better than twice and thrice aggregations. In the case of the same number of aggregation operations, we find that it is better to aggregate all nodes than top-$100$ nodes when using once and thrice aggregation operations. Based on the above observations, we adopt once aggregation operation and aggregate all superpoints to target superpoint during the spatial-structural diversity reasoning phase.

\textbf{Noise-aware Iterative Labeling.} We conduct several comparison experiments with different purity threshold $\theta$ in Figure~\ref{fig:analysis} (b). We observe that the mIoU on $\theta=0.9$ and $\theta=1.0$ are obviously better than $\theta=0.6$. Especially, the mIoU of $\theta=0.9$ is $2.73\%$ higher than $\theta = 0.6$ at the $2^{th}$ active learning cycle. Although in the first two active learning cycles, the mIoU of $\theta=1.0$ is 
lower than that of $\theta=0.9$, $\theta=1.0$ surpasses $\theta=0.9$ by a slight margin in the subsequent learning stage. In our opinion, the reason for this phenomenon is that $\theta=1.0$ discards more noisy superpoints and confounding areas than that of $\theta=0.9$ in the early cycles shown in Figure~\ref{fig:analysis} (c), which results in insufficient labeled data for training. But, it outperforms $\theta=0.9$ in the later active learning cycles since it gains enough training data while does not suffer from the negative effects of noisy superpoints.

\section{Conclusion}
In this paper, we target active learning for point cloud semantic segmentation and propose a novel superpoint-based SSDR-AL method. Through analyzing the spatial-geometrical information between superpoints, we utilize a weight-based superpoint uncertainty estimation and graph reasoning to select the most informative and representative superpoints. We further design a noise-aware iterative labeling strategy to accurately annotate clean areas and discard confounding areas of superpoints. We show the effectiveness of SSDR-AL on both S3DIS and Semantic3D datasets through systematic and comprehensive experiments. Our results strongly highlight the importance of using weight-based superpoint uncertainty estimation, diversity-based sampling strategy, and noise-aware iterative labeling strategy in such active learning methods in the future.

{\small
\bibliographystyle{ieee_fullname}
\bibliography{main}
}

\end{document}